# Instrumentality Tests Revisited

**Blai Bonet**
Cognitive Systems Laboratory
Department of Computer Science
University of California, Los Angeles
Los Angeles, CA 90024
bonet@cs.ucla.edu

## Abstract

An instrument is a random variable that is uncorrelated with certain (unobserved) error terms and, thus, allows the identification of structural parameters in linear models. In nonlinear models, instrumental variables are useful for deriving bounds on causal effects. Few years ago, Pearl introduced a necessary test for instruments which permits researchers to identify variables that could not serve as instruments. In this paper, we extend Pearl's result in several directions. In particular, we answer in the affirmative an open conjecture about the non-testability of instruments in models with unrestricted variables, and we devise new tests for models with discrete and continuous variables.

## 1 Introduction

Consider a simple structural equation $Y = \beta X + \varepsilon$ where $\varepsilon$ is a random variable with mean zero. We would like to estimate $\beta$, the causal effect of $X$ on $Y$, from a sample of $(X,Y)$ data. It is known that, absent of additional information about epsilon (e.g., that $\varepsilon$ and $X$ are uncorrelated) such estimation cannot be accomplished consistently. The method of instrumental variables is a way of integrating additional information in order to estimate $\beta$ consistently [11, 5]. An instrument $Z$ is a random variable that is correlated with $X$ and is judged to be uncorrelated with $\varepsilon$. Under such conditions the parameter $\beta$ becomes $E[ZY]/E[ZX]$ which can be estimated consistently from data, using ordinary least squares.

Thus, the problem of identifying the causal effect of $X$ on $Y$ becomes the problem of finding an appropriate instrument $Z$ that satisfies the conditions above. However, since these conditions involve an unobserved variable, $\varepsilon$, the selection of instruments has been a matter of judgment, unsupported by hard data. It is well known that no test for instruments exists when $X$, $Y$ and $Z$ are normally distributed; i.e., every tri-variate normal distribution is compatible with the assumption that $Z$ is uncorrelated with $\varepsilon$. Remarkably, this is not the case when the variables are discrete. Judea Pearl [7] derived a necessary test in the form of inequality that constrains the joint distribution whenever $Z$ is an instrument for $X$ and $Y$. (That is, whenever Z is independent of $\varepsilon$, and Y is some function of $X$ and $\varepsilon$). However, the existence of such a test for the general case of continuous variable has remained undecided. In this paper, we show that no such test exists when $X$ is continuous, and we further devise new tests for instrumental variables in the case where $X$ is discrete.

The paper is organized as follows. Section 2 presents a canonical form of the general model in which $X$ is a cause of $Y$ and $Z$ is an instrument for $X$ and $Y$. Section 3 derives properties of the set of probability distributions compatible with the general model, provides an alternative proof of the necessity of Pearl's test, and shows, by means of an example, that the test is not complete. Section 4 studies the asymptotic behavior of the properties derived in Section 3 and projects these properties onto the continuous case. Section 5 and 6 present stronger instrumentality tests for the discrete and continuous cases.

## 2 Canonical Model

The general problem can be represented by four random variables $X$, $Y$, $Z$ and $U$ such that $Z$ is independent of $U$ and $Y$ is conditionally independent of $Z$ given $X$ and $U$ (see [7]). The ***unique*** Bayesian network that satisfy this two conditions is shown in Fig. 1 (which we shall call from now on just the model). Here, the variable $U$ corresponds to the unobserved error term in the structural equation. It is known that any Bayesian network can be represented by another Bayesian network in which every interior node has a new parent (which is a root), and where all condi-tional probability tables (CPTs) represent functional relationships (see [2]). We will use that representation since it makes things easier. Observe that the Bayesian network of Fig. 1 satisfies the first condition, so we can assume without loss of generality that its CPTs represent functional relationships; i.e., that there exist functions $g$ and $h$ such that $X = g(Z, U)$ and $Y = h(X, U)$.

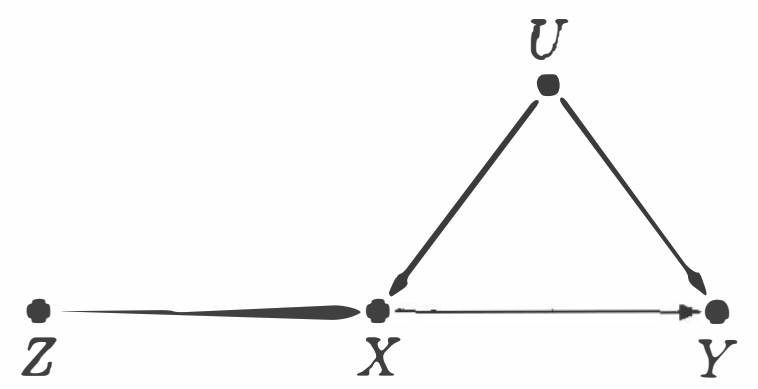

Figure 1: Graphical model for variables $X$, $Y$, $Z$ and $U$. This model satisfies the independences $(Z \perp\!\!\!\perp U \mid \emptyset)$ and $(Y \perp\!\!\!\perp Z \mid X, U)$. The variable $Z$ is called an instrument for the identification of the effect of $X$ on $Y$ since it is uncorrelated with $U$, and its effect on $Y$ is mediated by $X$.

The problem of testing whether $Z$ is a instrument for the effect of $X$ on $Y$ is analogous to the problem of determining which constraints are imposed by the structure of the Bayesian network on the set of all probability distributions compatible with it. A treatment of this problem for general Bayesian networks was given by Geiger and Meek [4]. Their approach, based on ideas from algebraic topology, represents the constraints as a surface in multi-dimensional Euclidean space. However, our inability to deal with such surface thwarts us from using their approach in this particular problem. Instead, we will use a simpler approach based on convex analysis to obtain some novel and interesting results; the bad news is that our method is not as general as Geiger and Meek's.

Let us start with some basic definitions and examples. Assume that $X$, $Y$ and $Z$ are discrete variables with *finite domains* $\mathcal{X} = \{x_1, \ldots, x_n\}$, $\mathcal{Y} = \{y_1, \ldots, y_m\}$ and $\mathcal{Z} = \{z_1, \ldots, z_l\}$, and with no restrictions on $\mathcal{U}$, the domain of $U$. Thus, the network can represents any finite discrete distribution $P(X, Y, Z)$ in which $Z$ is a instrument. In this case, Pearl showed that the following inequality is a necessary condition for the three variables being generated by the model

$$\max_{i=1\ldots n} \sum_{j=1}^{m} \max_{k=1\ldots l} P(x_i, y_j | z_k) \leq 1. \tag{1}$$

The condition above is equivalent to the set of inequalities that result after the $z_k$ for all possible values in $\mathcal{Z}$; i.e.,

$$\sum_{j=1}^{m} P(x_i, y_j | z_{k_{i,j}}) \leq 1 \tag{2}$$

for all $1 \leq i \leq n$, $k_{i,j} \in \{1, \ldots, l\}$. It is easy to see that there are $nl^m$ such equations. Necessity means that the researcher must consider $Z$ as a *candidate* instrument only if Eq.(1) is satisfied. For this reason, Eq.(1) was named the *instrumental inequality* by Pearl.

When all variables are binary, for example, there are 8 instrumental inequalities

$$\begin{aligned}
P(x_1, y_1|z_1) + P(x_1, y_2|z_1) &\leq 1 \\
P(x_1, y_1|z_1) + P(x_1, y_2|z_2) &\leq 1 \\
P(x_1, y_1|z_2) + P(x_1, y_2|z_1) &\leq 1 \\
P(x_1, y_1|z_2) + P(x_1, y_2|z_2) &\leq 1 \\
P(x_2, y_1|z_1) + P(x_2, y_2|z_1) &\leq 1 \\
P(x_2, y_1|z_1) + P(x_2, y_2|z_2) &\leq 1 \\
P(x_2, y_1|z_2) + P(x_2, y_2|z_1) &\leq 1 \\
P(x_2, y_1|z_2) + P(x_2, y_2|z_2) &\leq 1.
\end{aligned} \tag{3}$$

Of all these, only 4 are non-trivial (the 2nd, 3rd, 6th and 7th); i.e., there are probability distributions that do not satisfy them.

We associate to each probability distribution $P$ a $mnl$-dimensional real vector $F(P)$ defined as

$$F(P) \stackrel{\text{def}}{=} (P(x_1, y_1|z_1), P(x_1, y_2|z_1), \ldots, P(x_1, y_m|z_1), P(x_1, y_1|z_2), \ldots, P(x_n, y_m|z_l)).$$

Let $\mathsf{F}$ be the set of all such vectors and $\mathsf{T}$ the set of all $nl^m$-dimensional real vectors. It is not hard to see that the vector of left-hand side expressions in Eq.(2) is the image of a linear transformation $A_1 : \mathsf{F} \to \mathsf{T}$ where $A_1$ is a $nl^m \times mnl$ matrix of zeros and ones. Fig. 2, for example, shows the matrix $A_1$ for the binary case; the first row of $A_1$ corresponds to the first equation in (3), the second row to the second equation, and so on.

### 2.1 Response Variables

The $U$ variable can be interpreted as a "selector" of functions for the $X$ and $Y$ variables from the sets $\mathcal{G} = \{g : g \text{ is function } \mathcal{Z} \to \mathcal{X}\}$ and $\mathcal{H} = \{h : h \text{ is function } \mathcal{X} \to \mathcal{Y}\}$ respectively. Consider a fixed value $z \in \mathcal{Z}$, and let $x, y$ be two values for $X, Y$. Denote with $\mathcal{G}_{zx}$ and $\mathcal{H}_{xy}$ the sets $\{g \in \mathcal{G} : g(z) = x\}$ and $\{h \in \mathcal{H} : h(x) = y\}$ respectively. It is easy to see that $\#\mathcal{G}_{zx} = n^{l-1}$ and $\#\mathcal{H}_{xy} = m^{n-1}$. Let $r_{zx \cdot j}$ and $s_{xy \cdot j}$ denote the $i$th function from $\mathcal{G}_{zx}$ and the $j$th function from $\mathcal{H}_{xy}$ for $1 \leq i \leq n^{l-1}$ and $1 \leq j \leq m^{n-1}$. Note that $(r_{zx \cdot i}, s_{xy \cdot j}) = (r_{zx' \cdot i'}, s_{x'y' \cdot j'})$ if and only

$$A_1 = \left[\begin{array}{cccc|cccc} 1 & 1 & 0 & 0 & 0 & 0 & 0 & 0 \\ 1 & 0 & 0 & 0 & 0 & 1 & 0 & 0 \\ 0 & 1 & 0 & 0 & 1 & 0 & 0 & 0 \\ 0 & 0 & 0 & 0 & 1 & 1 & 0 & 0 \\ \hline 0 & 0 & 1 & 1 & 0 & 0 & 0 & 0 \\ 0 & 0 & 1 & 0 & 0 & 0 & 0 & 1 \\ 0 & 0 & 0 & 1 & 0 & 0 & 1 & 0 \\ 0 & 0 & 0 & 0 & 0 & 0 & 1 & 1 \end{array}\right], \quad A_2 = \left[\begin{array}{cccc|cccc|cccc|cccc} & \star & & \star & & & & & & & & & & & \star & \star \\ 1 & 1 & 0 & 0 & 1 & 1 & 0 & 0 & 0 & 0 & 0 & 0 & 0 & 0 & 0 & 0 \\ 0 & 0 & 0 & 0 & 0 & 0 & 0 & 0 & 1 & 0 & 1 & 0 & 1 & 0 & 1 & 0 \\ 0 & 0 & 1 & 1 & 0 & 0 & 1 & 1 & 0 & 0 & 0 & 0 & 0 & 0 & 0 & 0 \\ 0 & 0 & 0 & 0 & 0 & 0 & 0 & 0 & 0 & 1 & 0 & 1 & 0 & 1 & 0 & 1 \\ \hline 1 & 1 & 0 & 0 & 0 & 0 & 0 & 0 & 1 & 1 & 0 & 0 & 0 & 0 & 0 & 0 \\ 0 & 0 & 0 & 0 & 1 & 0 & 1 & 0 & 0 & 0 & 0 & 0 & 1 & 0 & 1 & 0 \\ 0 & 0 & 1 & 1 & 0 & 0 & 0 & 0 & 0 & 0 & 1 & 1 & 0 & 0 & 0 & 0 \\ 0 & 0 & 0 & 0 & 0 & 1 & 0 & 1 & 0 & 0 & 0 & 0 & 0 & 1 & 0 & 1 \end{array}\right]$$

Figure 2: Example of matrices $A_1$ and $A_2$ for the binary case. $A_1$ is a $nl^m = 8$ times $mnl = 8$ matrix and $A_2$ is a $mnl = 8$ times $n^l m^n = 16$ matrix. The $\star$ mark repeated columns.

| $\mathcal{Z} \to \mathcal{X}$ | $\mathcal{X} \to \mathcal{Y}$ |
|---|---|
| (0,0) | (0,0) |
| (0,1) | (0,1) |
| (1,0) | (1,0) |
| (1,1) | (1,1). |

Table 1: Functions that generate the partition $\mathbb{P}$ for the binary case. Each function is represented by a pair in $\mathcal{X}^2$ or $\mathcal{Y}^2$ depending whether it is in $\mathcal{Z} \to \mathcal{X}$ or $\mathcal{X} \to \mathcal{Y}$. The functions are ordered by row number; e.g., the third function from $\mathcal{Z} \to \mathcal{X}$, denoted by $(1,0)$, is given by $g(z_1) = x_2$ and $g(z_2) = x_1$.

if $(x, y, i, j) = (x', y', i', j')$, and that $(r_{zx \cdot i}, s_{xy \cdot j}) = (r_{z'x' \cdot i'}, s_{x'y' \cdot j'})$ for some $x', y', i', j'$. Thus, the collection

$$\mathbb{P} = \{ (r_{zx \cdot i}, s_{xy \cdot j}) \ : x \in \mathcal{X}, y \in \mathcal{Y}, \\ 1 \leq i \leq n^{l-1}, 1 \leq j \leq m^{n-1} \}$$

is a partition of $\mathcal{U}$ into $n^l m^n$ pieces. This partition, called response variables or mapping variables, has been used before to derive bounds for the causal effect of $X$ on $Y$ [1, 3]. When all variables are binary, there are 16 pairs in $\mathbb{P}$ that corresponds to the cross product of the sets of functions in Table 1.

The collection of probability distributions compatible with the model are those that can be generated when assigning probabilities to the pairs in $\mathbb{P}$. There are $n^l m^n$ such pairs, so each model can be represented by a $n^l m^n$-dimensional stochastic vector. It is easy to check that any conditional probability $P(xy|z)$ can be expressed by the sum

$$P(xy|z) = \sum_{i=1}^{n^{l-1}} \sum_{j=1}^{m^{n-1}} P(r_{zx.i}, s_{xy.j}). \qquad (4)$$

Thus, if Q is the set of all $n^l m^n$-dimensional stochastic vectors and $\mathsf{B} \subseteq \mathsf{F}$ the subset of F that corresponds to the model, then each $B \in \mathsf{B}$ is the image of some $Q \in \mathsf{Q}$ under a linear transformation $A_2 : \mathsf{Q} \to \mathsf{B}$, where $A_2$ is a $mnl \times n^l m^n$ matrix of zeros and ones. Likewise, the collection of vectors for the left-hand sides of Eq.(2) for the model is the image of Q under the composition of $A_1$ and $A_2$; i.e., $A_1 A_2 \mathsf{Q}$.

In the example, the probability $P(X = 0, Y = 0|Z = 0)$ is expressed as the sum

$$P(r_{00 \cdot 1}, s_{00 \cdot 1}) + P(r_{00 \cdot 1}, s_{00 \cdot 2}) + \\ P(r_{00 \cdot 2}, s_{00 \cdot 1}) + P(r_{00 \cdot 2}, s_{00 \cdot 2}).$$

Fig. 2 also shows the matrix $A_2$ in which the column $4(i-1)+j$ correspond to $i$th function from $\mathcal{Z} \to \mathcal{X}$ and $j$th function from $\mathcal{Z} \to \mathcal{X}$ for $i, j \in \{1, \ldots, 4\}$ (see Table 1). For example, the reader can check that the first row of $A_2$ corresponds to above expression.

## 3 Geometric Properties

In this section we show some convex properties of the sets Q, B and F, give an alternate proof of the necessity of Eq.(2), and show that the test is not complete; i.e., that there exists a probability distribution that satisfies the instrumental inequality but is not generated by the model. Some results of this section will be used later to prove properties in the continuous case.

**Lemma 1** *F is a polyhedral convex set with $(mn)^l$ extreme points.*

*Proof:* It is not hard to see that F is convex. The set of $(mn)^l$ points given by setting $P(x_{i_k}, y_{j_k}|z_k) = 1$ for $k = 1 \ldots l$ for all possible indexes $i_k$ and $j_k$ are extreme points. All points in F have coordinates in $[0,1]^{mnl}$. Using induction, it is easy to see that any $F \in \mathsf{F}$ can be expressed as a convex combination of above points. ■

**Lemma 2** *The matrix $A_3 \stackrel{def}{=} A_1 A_2$ is made of zeros and ones.*

*Proof:* Consider two arbitrary terms $P(xy|z)$ and $P(xy'|z')$ of an instrumental inequality. They share no response variable since

$$P(xy|z) = \sum_{i=1}^{n^{l-1}} \sum_{j=1}^{m^{n-1}} P(r_{zx\cdot i}, s_{xy\cdot j}),$$

$$P(xy'|z') = \sum_{i=1}^{n^{l-1}} \sum_{j=1}^{m^{n-1}} P(r_{z'x\cdot i}, s_{xy'\cdot j}),$$

$y \neq y'$, and $s_{xy\cdot j} \neq s_{xy'\cdot i}$ for all $i$ and $j$. ■

**Lemma 3** *Q is a polyhedral convex set with $n^l m^n$ extreme points, B is a polyhedral convex set, and each extreme point of B is an extreme point of F.*

*Proof:* Q is obviously a polyhedral convex set since it is the collection of probability distributions over $n^l m^n$ points. Each extreme point is a vector of $n^l m^n - 1$ zeros and 1 one, so Q has $n^l m^n$ extreme points. B is the image of a polyhedral convex set under the linear transformation $A_2$, so it is a polyhedral convex set. Each extreme point of B is the transformation of an extreme point of Q, so its coordinates are either 0 or 1. Since B $\subseteq$ F, then each extreme point of B is an extreme point of F. ■

**Theorem 4 (Necessity)** *The instrumental inequality is satisfied by all $P \in$ B.*

*Proof:* Consider an $nl^m$-dimensional real vector $T \in$ T generated by the model. Then, the following linear system, called Primal, must has solution

$$\begin{array}{lll} \sum_{i=1}^{n^l m^n} Q_i = 1 & & \text{(1 equation)} \\ A_3 Q = T & & (nl^m \text{ equations}) \\ Q_i \geq 0 & i = 1 \ldots n^l m^n & (n^l m^n \text{ equations}) \end{array}$$

where $Q_i$ denotes the $i$th coordinate of $Q$. It is a system of $1 + nl^m + n^l m^n$ equations in $n^l m^n$ unknows, and it is intimately related to another system of $n^l m^n$ inequalities in $nl^m + 1$ unknowns called Dual:

$$\pi_0 + \pi' A_3^j \leq 0 \qquad j = 1 \ldots n^l m^n \qquad (5)$$
$$\text{unrestricted } \pi_0, \pi$$

where $\pi_0$ is a scalar, $\pi$ is a $nl^m$-dimensional real vector, and $A_3^j$ is the $j$th column of $A_3$. The relation being that the former has solution if and only if the inequalities (5) imply the inequality

$$\pi_0 + \pi' T \leq 0. \qquad (6)$$

A fact that is true if and only if there exists $n^l m^n$ non-negative scalars $\lambda_1, \ldots, \lambda_{n^l m^n}$ such that (6) can be expressed as a combination of the inequalities in (5) and the $\lambda_i$'s (see Rockafellar [10]; Sect. 22). We proceed by reasoning with the $\lambda_i$'s. Since $\pi_0$ appears with coefficient 1 in all inequalities then $\sum_{j=1}^{n^l m^n} \lambda_j = 1$. Thus, all coordinates in $T$ must be less than or equal to 1 since the matrix $A_3$ has only zeros and ones (by Lemma 2). ■

If we replace the matrix $A_3$ by $A_2$, and $T$ by a vector $B \in$ B in the proof, then the relation between (5) and (6), and $\sum_i \lambda_i = 1$ imply

**Corollary 5** *The extreme points of B are the columns of matrix $A_2$.*

We use this fact to count the number of extreme points in B. This information will be used to see what happens when the size of the domains grows up to infinity.

**Theorem 6** *The set B has*

$$\sum_{k=1}^{n} \binom{n}{k} (-1)^k m^k \sum_{j=0}^{k} \binom{k}{j} (-1)^j j^l$$

*extreme points.*

*Proof:* Remember that each column of $A_2$ corresponds to a pair in $\mathcal{G} \times \mathcal{H}$. Consider two such pairs $p_1 = (g, h)$ and $p_2 = (g', h')$, and denote by col$(p_1)$ and col$(p_2)$ the corresponding columns. First, we prove that col$(g,h)$ = col$(g',h')$ if and only if $g = g'$ and $\{x : h(x) \neq h'(x)\} \cap g(\mathcal{Z}) = \emptyset$. If $g \neq g'$, then there exists $z$ such that $g(z) \neq g'(z)$ which implies the row for $P(g(z), h(g(z))|z)$ has a 1 in col$(g,h)$ and 0 in col$(g',h')$; i.e., they are different. Similarly, when $g = g'$ and there is a $z$ such that $h(g(z)) \neq h'(g(z))$. Now, suppose that col$(g,h) \neq$ col$(g',h')$. Then, there is row $P(x,y|z)$ such that col$(g,h)$ has 1 and col$(g',h')$ has 0. Thus, $g(z) = x$, $h(g(z)) = y$ and either $g'(z) \neq x$ or $h'(g(z)) \neq y$.

Now, partition $\mathcal{G}$ into $\{\mathcal{G}_k : k = 1 \ldots n\}$ such that $g \in \mathcal{G}_k$ if and only if $\#g(\mathcal{Z}) = k$. Fix $g \in \mathcal{G}_k$ and $h, h' \in \mathcal{H}$ such that col$(g,h)$ = col$(g,h')$. Then, by the first result, $\forall x[x \in g(\mathcal{Z}) \Rightarrow h(x) = h'(x)]$, so $\#\{h' \in \mathcal{H} : \text{col}(g,h) = \text{col}(g,h')\} = m^{n-k}$. Therefore, the number of distinct columns in $A_2$ is

$$\sum_{k=1}^{n} \frac{m^n}{m^{n-k}} \#\mathcal{G}_k = \sum_{k=1}^{n} m^k \#\mathcal{G}_k. \qquad (7)$$

We finish by computing the value $\#\mathcal{G}_k$. Fix $k$ elements $A = \{x_{i_1}, \ldots, x_{i_k}\} \subseteq \mathcal{X}$. By the principle of inclusion/exclusion applied to the sets of functions $\mathcal{Z} \to A, \mathcal{Z} \to A_{k-1}, \mathcal{Z} \to A_{k-1}, \ldots$ such that $A_q \subseteq A$ and $\#A_q = q$, we see that

$$\#\mathcal{G}_k = \binom{n}{k} \left[ k^l - \binom{k}{1} (k-1)^l + \cdots + 0 \right]$$

$$= \binom{n}{k} \sum_{j=0}^{k} \binom{k}{j} (-1)^j (k-j)^l$$
$$= \binom{n}{k} \sum_{j=0}^{k} \binom{k}{j} (-1)^{k-j} j^l.$$

The result is obtained by plugging this value in (7). ■

### 3.1 Incompleteness

Since the number of extreme points in B grows much faster than the number of inequalities in Eq.(2), there is no hope that the instrumental inequality will be a sufficient test for a probability distribution being generated by the model. A concrete example is given next for the case $n = m = 2$ and $l = 3$. Let $F_1, F_2 \in \mathsf{F}$ be given by

$$F_1 = (0,0,1,0,\ \ 0,0,1,0,\ \ 1,0,0,0)',$$
$$F_2 = (1,0,0,0,\ \ 0,1,0,0,\ \ 0,0,0,1)'.$$

They generate the points $T_1, T_2 \in \mathsf{T}$, and the convex combination $\widehat{T} = \alpha T_1 + (1-\alpha)T_2 \in \mathsf{T}$:

$$T_1 = (1,2,1,\ \ 0,1,0,\ \ 0,1,0,\ \ 0,0,1,\ \ 0,0,1,\ \ 0,0,1)',$$
$$T_2 = (0,0,0,\ \ 0,0,0,\ \ 1,1,1,\ \ 1,1,1,\ \ 1,1,1,\ \ 0,0,0)',$$
$$\widehat{T} = (\alpha,2\alpha,\alpha,\ \ 0,\alpha,0,\ \ 1-\alpha,1,1-\alpha,\ \ 1-\alpha,1-\alpha,1,\ \ 1-\alpha,1-\alpha,1,\ \ 0,0,\alpha)',$$

$\widehat{T}$ satisfies the instrumental inequality when $\alpha \leq 1/2$ but it can be generated by the model only if $\alpha = 0$.[1]

## 4 Asymptotic Analysis

It is known that when restricted to Gaussians the model in Fig. 1 imposes no constraints on the distributions; i.e., any tri-variate Gaussian is compatible with it (see [7]). This result plus the elusiveness of the model when $X$ is continuous caused some researchers to conjecture that the model imposes no constraints in such case. In this section, we prove that this conjecture is true when $Z$ and $Y$ are discrete of finite domain, or when $Y$ is continuous and $X$ and $Z$ are discrete of finite domain. We also show that the model is quite restrictive in other cases. We begin by studying the limiting behavior of the ratio between the number of extreme points of the set of distribution compatible with the model and the number of extreme points for general distributions; i.e., the quantity $R(l,m,n) \stackrel{\text{def}}{=} \#\text{ext}(\mathsf{B})/\#\text{ext}(\mathsf{F})$.

[1] This is not easy to check by hand. We have checked it with *Mathematica*.

Substituting the values given by Lemma 1 and Theorem 6, we get the expression

$$R(l,m,n) = \frac{1}{m^l n^l} \sum_{k=1}^{n} \binom{n}{k} (-1)^k m^k \sum_{j=0}^{k} \binom{k}{j} (-1)^j j^l.$$

It can be shown, using induction, that the inner sum is equal to $(-1)^k(\Delta^k x^l)(0)$ where $(\Delta^k x^l)(\cdot)$ is the $k$th-fold composition of the difference operator $\Delta$ over the function $f(x) = x^l$ (see [6]; pp. 187–188); i.e.,

$$\Delta^k f(x) = \begin{cases} f(x) & \text{if } k = 0, \\ \Delta^{k-1} f(x+1) - \Delta^{k-1} f(x) & \text{if } k \neq 0. \end{cases}$$

Since $(\Delta^k x^l) = 0$ for $k > l$ and $n \wedge l \stackrel{\text{def}}{=} \min\{n, l\}$, we have

$$\begin{aligned} \#\text{ext}(\mathsf{B}) &= \sum_{k=1}^{n} \binom{n}{k} m^k (\Delta^k x^l)(0) \\ &= \sum_{k=1}^{n\wedge l} \binom{n}{k} m^k (\Delta^k x^l)(0). \end{aligned}$$

In the binary case, for example,

$$\begin{aligned} \#\text{ext}(\mathsf{B}) &= \sum_{k=1}^{2} \binom{2}{k} 2^k (\Delta^k x^2)(0) \\ &= 2 \cdot 2 \cdot (\Delta^1 x^2)(0) + 2^2 \cdot (\Delta^2 x^2)(0) \\ &= 4[(x^2)(1) - (x^2)(0)] + \\ &\quad\ 4[(\Delta^1 x^2)(1) - (\Delta^1 x^2)(0)] \\ &= 4 + 4[(x^2)(2) - (x^2)(1) - 1] \\ &= 4 + 4[2^2 - 1 - 1] \\ &= 12 \end{aligned}$$

that is equal to the number of different columns in the matrix $A_2$ of Fig. 2.

We now see what happens when $l, m, n$ go to infinity independently. Fix $l, m$ and let $n \gg l$. Then,

$$\begin{aligned} R(l,m,n) &= \frac{1}{n^l m^l} \sum_{k=1}^{l} \binom{n}{k} m^k (\Delta^k x^l)(0) \\ &= \frac{1}{n^l m^l} \left[ \frac{n!}{(n-l)!\,l!} m^l l! + \right. \\ &\qquad \left. \sum_{k=1}^{l-1} m^k\, \Theta(n^k)\, (\Delta^k x^l)(0) \right] \\ &= \frac{1}{n^l m^l} \left[ n(n-1)\cdots(n-l+1) m^l + \right. \\ &\qquad \left. \sum_{k=1}^{l-1} m^k\, \Theta(n^k)\, \Theta(k^l) \right] \end{aligned}$$

$$= \frac{1}{n^l m^l}\left[ n^l m^l + m^l \Theta(n^{l-1}) + \sum_{k=1}^{l-1} m^k\, \Theta(n^k k^l) \right]$$
$$= 1 + \Theta(n^{-1}) + \Theta(n^{-1}m^{-1}(l-1)^l)$$

using the facts $(\Delta^l x^l)(0) = l!$ for all non-negative integers $l$, $\binom{n}{k} = \Theta(n^k)$ and $(\Delta^k x^l)(0) = \Theta(k^l)$. Thus, $R(l,m,n) \to 1$ as $n \to \infty$ for fixed $l$ and $m$.

Fix now $n, m$ and let $l \gg n$. Then

$$R(l,m,n) = \frac{1}{n^l m^l} \sum_{k=1}^{n} \binom{n}{k} m^k (\Delta^k x^l)(0)$$
$$= \frac{1}{n^l m^l} \sum_{k=1}^{n} \binom{n}{k} m^k\, \Theta(k^l)$$
$$= \frac{1}{n^l m^l} \Theta(n^n)\, \Theta(m^n)\, \Theta(n^l)$$
$$= \Theta(n^n m^{n-l}).$$

Thus, $R(l,m,n) \to 0$ as $l \to \infty$ for fixed $n$ and $m$. Finally, fix $l, n$ and suppose $m$ is big. Then,

$$R(l,m,n) = \frac{1}{n^l m^l} \sum_{k=1}^{n\wedge l} \binom{n}{k} m^k (\Delta^k x^l)(0)$$
$$= \frac{1}{n^l m^l} \sum_{k=1}^{n\wedge l} m^k\, \Theta(n^k)\, \Theta(k^l)$$
$$= \frac{1}{n^l m^l} \Theta(m^{n\wedge l})\, \Theta(n^{n\wedge l})\, \Theta((n \wedge l)^l).$$

If $n \geq l$, then $R(l,m,n) = \Theta(l^l)$. If $n < l$, then $R(l,m,n) = \Theta(m^{n-l} n^n)$. Thus, $R(l,m,n)$ goes to 0 or a constant as $m$ goes to infinity when $n \geq l$ or $n < l$ respectively. In summary,

$$\lim_{n\to\infty} R(l,m,n) = 1, \tag{8}$$
$$\lim_{l\to\infty} R(l,m,n) = 0, \tag{9}$$
$$\lim_{m\to\infty} R(l,m,n) = \begin{cases} \Theta(1) & \text{if } n \geq l, \\ 0 & \text{if } n < l. \end{cases} \tag{10}$$

Remember that each extreme point of B is also an extreme point of F (Lemma 3). Therefore, Eq.(9) says that when $Z$ is a variable with infinite domain and $X, Y$ are variables of finite domain, then the model is "very" restrictive. Likewise, when $l \leq n < \infty$ and $Y$ is a variable with infinite domain. Eq.(8), on the other hand, *suggests* that when $X$ is a random variable of infinite domain (discrete or continuous) and $Y, Z$ are variables of finite domain, then the model imposes no constraints. In the rest of the section we sketch a formal proof of this claim. That is, we answer in affirmative the conjecture about that the model imposes no constraints when $X$ is continuous.

Assume that $Y$ and $Z$ are finite discrete variables with domains of size $m$ and $l$ respectively, and $X$ is a discrete variable of infinite domain. Let $\{X_n\}$ be a sequence random variables with distributions $P_n$ such that: *(i)* the domain of $X_n$ is finite of size $n$, and *(ii)* $P_n \to P$ (this is always possible). Since each probability $P_n(x, y|z)$ is the sum of $n^{l-1}m^{n-1}$ terms (see Eq.(4)), then for a sufficiently large $n$ the equation $A_2 Q = F(P_n)$ has solution in Q. Indeed, if $n$ is such that $n^{l-1}m^{n-1} > mnl$, then $P_{n'}$ is compatible with the model for all $n' > n$. Thus, $P$ is compatible with the model. For the case of continuous $X$, let $\{X_n\}$ be a sequence of discrete random variables such that $X_n \to X$ (always possible). Then, $P$ is compatible with the model since each $P_n$ is compatible.

## 5 Stronger Tests

In this section we show some results for more general tests and present a new stronger test. Since the extreme points of B are the columns of matrix $A_2$, then a necessary and sufficient test for a distribution $P$ being compatible with the model is that $F(P)$ belongs to the convex set B. However, Corollary 5 tells us that such test is not practical since the number of different columns in $A_2$ is exponential. Thus, we need to look for other more economical tests.

Let $\mathsf{B}(l,m,n)$ and $\mathsf{F}(l,m,n)$ denote the corresponding B and F sets for parameters $l$, $m$ and $n$ respectively. We make the following general definitions. A *$(l,m,n)$-test*, for set $F(l,m,n)$, is a pair $(\tau, \alpha)$ where $\tau$ is $mnl$-dimensional vector of non-negative integers and $\alpha$ is a positive integer. A vector $F \in \mathsf{F}(l,m,n)$ is said to ***pass*** $(\tau,\alpha)$ iff $\tau' F \leq \alpha$. If $(\tau,\alpha)$ is such that $F \in \mathsf{B}(l,m,n) \Rightarrow \tau' F \leq \alpha$, then we say that the test is *necessary* for $\mathsf{B}(l,m,n)$. If it is such that $\tau' F \leq \alpha \Rightarrow F \in \mathsf{B}(l,m,n)$, then we say that the test is *sufficient* for $\mathsf{B}(l,m,n)$. The useful tests are the necessary ones since they allow the researcher to discard potential instruments.

A $(l',m',n')$-test $(\tilde\tau, \alpha)$ is an *extension* of test $(\tau,\alpha)$, for $l' \geq l$, $m' \geq m$ and $n' \geq n$, if the "formal" expressions $\tau' F$ and $\tilde\tau' \tilde F$ are identical for $F \in \mathsf{F}(l,m,n)$ and $\tilde F \in \mathsf{F}(l',m',n')$. A necessary test $(\tau,\alpha)$ is called *regular* if, for every permutation of the domains of the variables, the test $(\tilde\tau,\alpha)$ is also necessary, where $\tilde\tau$ is the result of applying the permutation to $\tau$. In that case, the test $(\tilde\tau,\alpha)$ is called a *regular variation* of $(\tau,\alpha)$.

For example, Pearl's instrumental inequality is the col-lection of tests $\{(\tau_j, 1) : j = 1 \ldots nl^m\}$ where $\tau_j$ is the $j$th row of the matrix $A_1$. In this case, the instrumental inequality can be described by the two tests

$$((1,1,0,0,0,0,0,0),1)$$
$$((1,0,0,0,0,1,0,0),1);$$

i.e., by the tests

$$P(x_1,y_1|z_1) + P(x_1,y_2|z_1) \le 1$$
$$P(x_1,y_1|z_1) + P(x_1,y_2|z_2) \le 1$$

and all their regular variations. We have the following result

**Theorem 7** *Let $(\tau, \alpha)$ be a necessary instrumental test for $\mathsf{B}(l,m,n)$. Then, it is regular and all its extensions are necessary.*

*Proof (Sketch):* This is one of those theorems in which writing the proof is more difficult than the theorem itself. That any extension is also necessary should be obvious. That it is regular follows from the fact that there is nothing special about some particular value for the variables; i.e., the model is symmetric with respect to the values of the variable. ∎

In order to derive new instrumental tests, note that there are two standard representations of polyhedral convex sets: by enumerating its extreme points, or by a set of inequalities defining the 'faces' of the set. In our case, the first representation is given by the columns of $A_2$ and a representation of the second kind would correspond to a necessary and sufficient test. Researchers from the University of Heidelberg have developed a computer program called PORTA [8] that allows to go from (any) one of the representations to the other. We have used this program to check that Pearl's instrumental inequality is a *sufficient* test for the sets $\mathsf{B}(2,2,2)$, $\mathsf{B}(2,2,3)$ and $\mathsf{B}(2,3,2)$. However, it is not sufficient for $\mathsf{B}(3,2,2)$ since it needs to be completed with

$$P(x_1,y_2|z_2) + P(x_1,y_1|z_3) + P(x_1,y_2|z_1) + P(x_2,y_2|z_2) + P(x_2,y_1|z_1) \le 2. \quad (11)$$

and all its regular variations. Eq.(11) is a stronger version of Eq.(2) since it is the sum of two instrumental inequalities plus a non-negative quantity. By Theorem 7, the regular test corresponding to Eq.(11) is a stronger necessary test for $\mathsf{B}(l,m,n)$ when $l \ge 3$. Using this method, we have derived other stronger tests but them cannot be described nicely. Interestingly, Eq.(11) has a nice connection with a special graph. We wonder if this kind of connection can be generalized to other stronger tests.

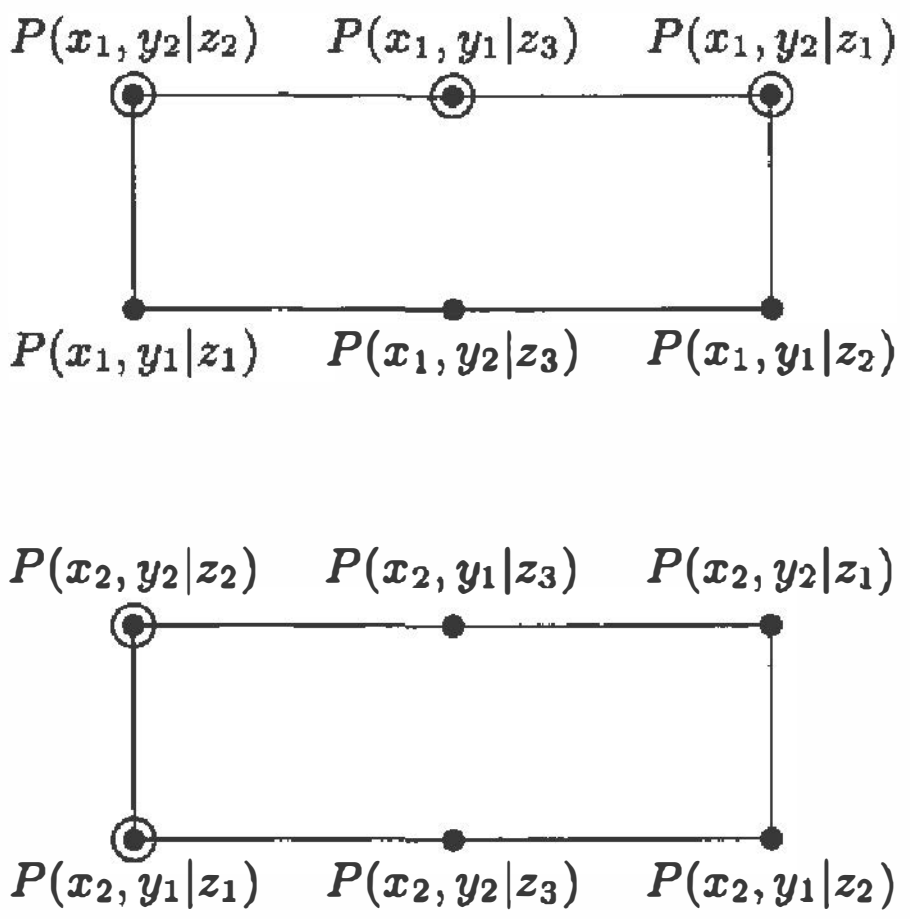

Figure 3: Graph that generates all regular variations of Eq.(11).

Consider the graph in Fig. 3. Observe that the terms in Eq.(11) correspond to the double circled nodes in the graph. It can be shown that all regular variations of Eq.(11) correspond to the inequalities whose left-hand side terms are obtained by picking five nodes in that graph subject to the following two restrictions:

*(i)* there should be three nodes in one component of the graph and two nodes in the other component, and

*(ii)* if we add edges between terms that only differ in the name for $x$ (e.g., between the nodes $P(x_1,y_1|z_1)$ and $P(x_2,y_1|z_1)$), then the subgraph induced by the chosen nodes is a tree (i.e., acyclic connected graph).

Check the example in Fig. 3. Note that if $n > 2$ then all possible pairs of values of $X$ need to be considered. Thus, for example, if $n = 3$, then all regular variations are obtained from the different graphs for $(x_1,x_2)$, $(x_1,x_3)$ and $(x_2,x_3)$.

## 6 The Case of Continuous $Z$ and $Y$

Now, we consider the case when $X$ is a discrete random variable (finite or infinite) and $Y, Z$ are continuous random variables. Let $\mu$ be the probability distribution for $U$, and $P(\cdot|x,u), P(\cdot|z,u)$ conditional probabilities for $Y, X$ respectively.[2] We will derive a necessary condition in this case that is very close to the instrumental inequality. The proof is inspired by Pearl's proof for the discrete case.

Fix $\varepsilon > 0$, a value $x$ for $X$, and let $\{B_i : i \ge 1\}$ be a

[2]Strictly speaking, they need to be regular conditional probabilities.

Borel partition of the real line; i.e., a partition of the line such that each $B_i$ is a Borel set. Note that the $B_i$'s could be intervals or more complex sets. By the definition of supremum, we can choose values $z(i,x)$ for $Z$ such that

$$P(Y \in B_i, x | z(i,x)) \geq \sup_z P(Y \in B_i, x|z) - \frac{\varepsilon}{2^i}$$

Then,

$$\begin{aligned}
&\sum_{i\geq 1} \sup_z P(Y \in B_i, x|z) \\
&\quad \leq \sum_{i\geq 1} \left[ P(Y \in B_i, x|z(i,x)) + \frac{\varepsilon}{2^i} \right] \\
&\quad = \sum_{i\geq 1} P(Y \in B_i, x|z(i,x)) + \varepsilon \\
&\quad = \sum_{i\geq 1} \int P(Y \in B_i|x,u) P(x|z(i,x),u) \mu(du) + \varepsilon \\
&\quad = \sum_{i\geq 1} E\Big[ P(Y \in B_i|x,U) P(x|z(i,x),U) \Big] + \varepsilon \\
&\quad = E\Big[ \sum_{i\geq 1} P(Y \in B_i|x,U) P(x|z(i,x),U) \Big] + \varepsilon \\
&\quad \leq 1 + \varepsilon\,.
\end{aligned}$$

Last equality by Lebesgue's monotone convergence theorem, and the last inequality since the interior sum is a convex sum of numbers bounded by 1. Now, let $\varepsilon \downarrow 0$ and take the sup over $\mathcal{X}$ to get the proof of

**Theorem 8** *Let $X$, $Y$ and $Z$ be random variables compatible with the model in Fig. 1 such that $X$ is discrete. Let $\{B_i : i \geq 1\}$ be a Borel partition of the real line. Then,*

$$\sup_x \sum_{i\geq 1} \sup_z P(Y \in B_i, x|z) \leq 1.$$

## 7 Summary

This paper derives new properties of the general, non-parametric model of instrumental variables and devises new tests for instrumentality. Combining tools from convex analysis, combinatorics and probability theory, enabled us to give an alternate proof of Pearl's test, show that the test is not complete, and settle an open conjecture about (the absence of) testable implications of an instrumental model when $X$ is a continuous variables. Other results are a general method for deriving stronger instrumental tests for the case of discrete variables (in particular the test of Eq.(11)), and a test for the case of discrete $X$ and continuous $Y$ and $Z$.

## Acknowledgements

I have had valuable discussions with Jin Tian and Carlos Brito. Judea Pearl is responsible of my interest in the topic; he also pointed me to improvements, related work, and some unsolved questions. Thomas Liggett from UCLA Math Department helped me clarify subtle issues associated with the continuous case. Finally, the comments from the UAI reviewers spotted a number of errors and helped me improve the paper. Many thanks to all of them.

## References

[1] A. Balke and J. Pearl. Probabilistic evaluation of counterfactual queries. In *Proc. of AAAI*, Volume I, pages 230–237. MIT Press, Menlo Park, CA, 1994.

[2] M.J. Druzdzel and H.A. Simon. Causality in Bayesian belief networks. In *Proc. of UAI*, Volume 9, pages 3–11. San Mateo, CA. Morgan Kaufmann, 1993.

[3] D. Heckerman and R. Shachter. Decision-theoretic foundations for causal reasoning. *Journal of Artificial Intelligence Research*, Volume 3, pages 405–430, 1995.

[4] D. Geiger and C. Meek. Graphical models and exponential families. In *Proc. of UAI*, Volume 14, pages 156–165. San Francisco, CA. Morgan Kaufmann, 1998.

[5] A.S. Goldberger. Structural Equation Models in the Social Sciences. In *Econometrica*, Volume 40, pages 979–1001, 1972.

[6] R. Graham and D. Knuth and O. Patashnik. Concrete Mathematics. First Edition. Addison-Wesley Publishing Company, 1989.

[7] J. Pearl. On the testability of causal models with latent and instrumental variables. In *Proc. of UAI*, Volume 11, pages 435–443. Morgan Kaufmann, 1995.

[8] PORTA. A Polyhedron Representation Transformation Algorithm. `http://www.iwr.uni-heidelberg.de/groups/comopt`

[9] J. Pearl. Causality: Models, Reasoning, and Inference. Cambridge University Press, 2000.

[10] T. Rockafellar. Convex Analysis. Princeton Landmarks in Mathematics and Physics, Princeton University Press, 1970, reprinted 1997.

[11] P.G. Wright. The Tariff on Animals and Vegetable Oil. McMillan, New York, 1928.